\pdfoutput=1

\documentclass[11pt]{article}

\usepackage{acl}

\usepackage{times}
\usepackage{latexsym}
\usepackage{float}

\usepackage[T1]{fontenc}

\usepackage[utf8]{inputenc}

\usepackage{microtype}
\usepackage{amsmath, amssymb}
\usepackage{booktabs}
\usepackage{graphicx}
\usepackage{multirow}

\title{Hierarchical Curriculum Learning for AMR Parsing}

\author{
Peiyi Wang$^1$$\footnotemark[1]$, \ Liang Chen$^1$$\footnotemark[1]$, \  Tianyu Liu$^2$, \  Damai Dai$^1$, \\
\textbf{Yunbo Cao$^2$, \  Baobao Chang$^1$, \  Zhifang Sui$^1$}
\\ 
$^1$ Key Laboratory of Computational Linguistics, Peking University, MOE, China \\
$^2$ Tencent Cloud Xiaowei \\
 \texttt{wangpeiyi9979@gmail.com; leo.liang.chen@outlook.com} \\
 \texttt{\{rogertyliu, yunbocao\}@tencent.com} \\
 \texttt{\{daidamai, chbb, szf\}@pku.edu.cn}
}

\date{}

\begin{document}

\maketitle

\renewcommand{\thefootnote}{\fnsymbol{footnote}}
\begin{abstract}
Abstract Meaning Representation (AMR) parsing aims to translate sentences to semantic representation with a hierarchical structure, and is recently empowered by pretrained sequence-to-sequence models. However, there exists a gap between their flat training objective (i.e., equally treats all output tokens) and the hierarchical AMR structure, which limits the model generalization. To bridge this gap, we propose a Hierarchical Curriculum Learning (HCL) framework with Structure-level (SC) and Instance-level Curricula (IC). SC switches progressively from core to detail AMR semantic elements while IC transits from structure-simple to -complex AMR instances during training. Through these two warming-up processes, HCL reduces the difficulty of learning complex structures, thus the flat model can better adapt to the AMR hierarchy. Extensive experiments on AMR2.0, AMR3.0, structure-complex and out-of-distribution situations verify the effectiveness of HCL.
\end{abstract}
\footnotetext[1]{Equal Contribution.}
\renewcommand{\thefootnote}{\arabic{footnote}}

\section{Introduction}
\begin{figure}[t]
    \centering
    \includegraphics[width=0.95\linewidth]{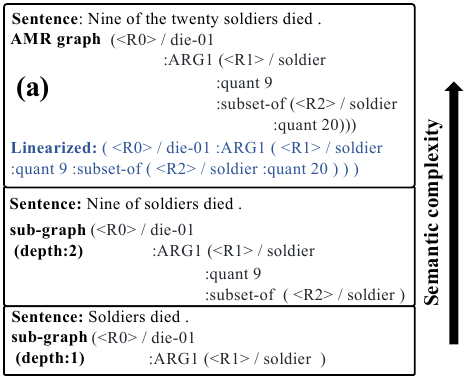}
    \caption{The AMR (sub-)graphs of the sentence ``Nine of the twenty soldiers died''. The deeper sub-graphs contain more sophisticated semantics compared with shallower ones. }
    \label{fig:example}
\end{figure}

Abstract Meaning Representation (AMR) \cite{ban-AMR} parsing aims to translate a natural sentence into a directed acyclic graph.
Figure \ref{fig:example}(a) illustrates an AMR graph where nodes represent concepts, e.g., `die-01' and `soldier',  and edges represent relations, e.g., `:ARG1' and `:quant'.
AMR has been exploited in the downstream NLP tasks, including information extraction \cite{rao-amr-ie, wang-amr-ie, zhang-amr-ie}, text summarization \cite{liao-amr-tm, hardy-amr-tm} and question answering \cite{mitra-amr-qa, sacha-amr-qa}.

The powerful pretrained encoder-decoder models, e.g., BART \cite{lew-bart}, have been successfully adapted to the AMR parsing and became the mainstream and state-of-the-art methods  \cite{bevil-spring}.
Through directly generating the linearized AMR graph (e.g., Figure \ref{fig:example}(a)) from the sentence, these sequence-to-sequence methods \cite{xu-seqpretrain,bevil-spring} circumvent the complex data processing pipeline and can be easily optimized compared with transition-based or graph-based methods \cite{naseem2019rewarding, lee2020pushing,lyu2018amr, zhang2019amr, zhang2019broad, cai2020amr, zhou2021structure}.
However, there exists a gap between the flat sentence-to-AMR training objective\footnote{Flat means the objective equally treats all output tokens.} and AMR graphs, since sequence-to-sequence models deviate from the essence of graph representation.
Therefore, it is difficult for sequential generators to learn the inherent hierarchical structure of AMR \cite{zhou2021structure}.

\begin{figure*}[h]
    \centering
    \includegraphics[width=1\linewidth]{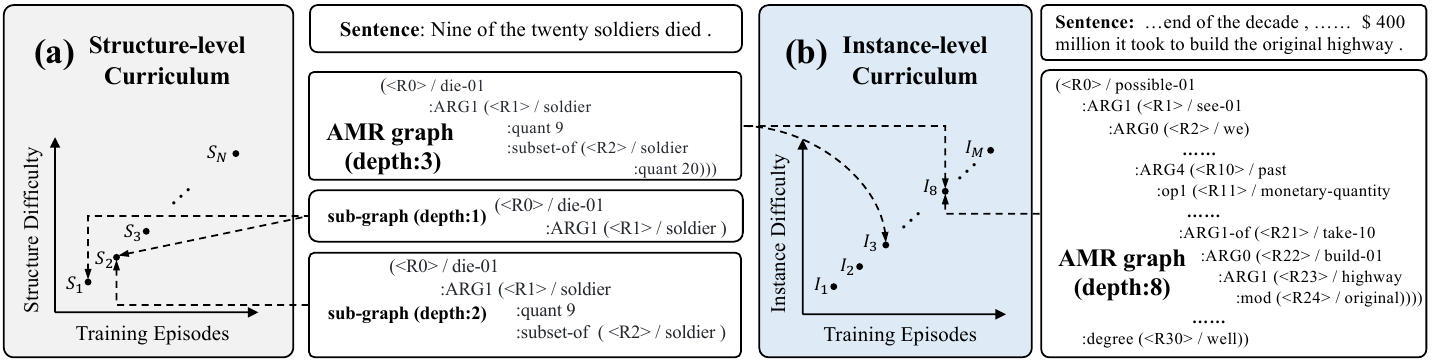}
    \caption{The overview of our hierarchical curriculum learning framework with two curricula, Structure-level (SC) and Instance-level Curricula (IC). During training, SC follows the principle of \textit{learning core semantics first}, which switches progressively from shallow to deep AMR sub-graphs. 
    IC follows the human intuition to \textit{start with easy instances}, which transits from easy to hard AMR instances.}
    \label{fig:overview}
\end{figure*}
\begin{figure}[t]
    \centering
    \includegraphics[width=0.9\linewidth]{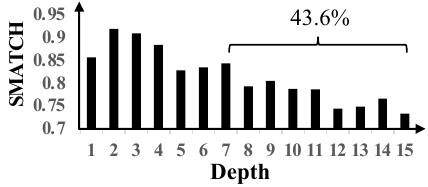}
    \caption{The average \textsc{smatch} scores for AMR graphs with different depths. The AMR graphs with at least depth $7$ accounted for $43.6$\% in the AMR-2.0 test set.}
    \label{fig:smatch-layer}
\end{figure}

Humans usually adapt to difficult tasks by dealing with examples gradually from easy to hard, i.e., Curriculum Learning \cite{bengio2009curriculum, platanios2019competence, su2020dialogue, xu-etal-2020-curriculum}.
Inspired by human behavior, we propose a hierarchical curriculum learning framework with two curricular strategies to help the flat pretrained model progressively adapt to the hierarchical AMR graph.
\textbf{(1) Structure-level Curriculum (SC)}.
AMR graphs are organized in a hierarchy where the core semantic elements stay closely to the root node \cite{cai-top2down}.
As depicted in Figure \ref{fig:example},
the concepts and relations that locate in the different layers of the AMR graph correspond to different levels of abstraction in terms of the semantic representation.
Motivated by the human learning process, i.e., \textbf{\textit{core concepts first, then details}},
SC enumerates all AMR sub-graphs with different depths, and deals with them in order from shallow to deep. 
\textbf{(2) Instance-level Curriculum (IC)}.
Our preliminary study in Figure \ref{fig:smatch-layer} shows that the performance of the vanilla BART baseline would drop rapidly as the depth of AMR graph grows, which indicates that handing deeper AMR hierarchy is more difficult for pretrained models.  
Inspired by the human cognition, i.e., \textbf{\textit{easy ones first, then hard ones}},
we propose IC which trains the model by starting from easy instances with a shallower AMR structure and then handling hard instances.

To sum up:
(1) Inspired by the human learning process, i.e., \textit{core concepts first} and \textit{easy instances first}, we propose a hierarchical curriculum learning (HCL) framework to help the sequence-to-sequence model progressively adapt to the AMR hierarchy.
(2) Extensive experiments on AMR2.0, AMR3.0, structure-complex and out-of-distribution situations verify the effectiveness of HCL.

\section{Methodology}
We formulate AMR parsing as a sequence-to-sequence transformation.
Given a sentence $\mathbf{x} = (x_1, ..., x_N)$, the model aims to generate a linearized AMR graph $\mathbf{y}=(y_1, ..., y_M)$.
As shown in Figure \ref{fig:example}(a),
following \citet{bevil-spring}, the AMR graph is linearized by the DFS-based linearization method with special tokens to indicate variables and parentheses to mark visit depth. 
Specifically, variables of AMR nodes are set to a series of special tokens \textit{<R0>, ..., <Rk>} (more details of linearization are included in Appendix \ref{apx:linearzation}).
In this paper, we propose a hierarchical curriculum learning framework (Figure \ref{fig:overview}) with the structure- and instance-level curricula to help the flat model progressively adapt to the structured AMR graph.

\subsection{Structure-level Curriculum}
\begin{table*}
    \centering
    \footnotesize
    \begin{tabular}{llccccccccc}
        \toprule
                ~& \multirow{2}{*}{Model} & \multirow{2}{*}{\textsc{smatch}} & \multicolumn{5}{c}{Structure-independent} & \multicolumn{3}{c}{Structure-dependent} \\
                \cmidrule(r){4-8} \cmidrule(r){9-11} 
        ~ & ~ & ~ & NoWSD & Conc. & NER & Neg. & Wiki. & Unll. & Reen. & SRL \\
        \midrule
        \multirow{8}{*}{\rotatebox[origin=c]{90}{{AMR2.0}}} & \citet{lyu2018amr}$^G$        
                                          & 74.4  & 75.5 & 85.9 & 86.0 & 58.4 & 75.7 & 77.1 & 52.3 & 69.8 \\
        ~& \citet{zhang2019amr}$^G$           & 76.3  & 76.8 & 84.8 & 77.9 & 75.2 & 85.8 & 79.0 & 60.0 & 69.7 \\
        ~& \citet{cai2020amr}          & 78.7  & 79.2 & 88.1 & 87.1 & 66.1 & 81.3 & 81.5 & 63.8 & 74.5 \\
        ~& \citet{cai2020amr}$^G$          & 80.2  & 80.8 & 88.1 & 81.1 & \textbf{78.9} & \textbf{86.3} & 82.8 & 64.6 & 74.2 \\
        ~& \citet{astudillo2020transition} & 80.2  & 80.7 & 88.1 & 87.5 & 64.5 & 78.8 & 84.2 & 70.3 & 78.2 \\
        ~& \citet{zhou2021amr} & 81.7  & 82.3 & 88.7 & 88.5 & 69.7 & 78.8 & 85.5 & 71.1 & 80.8 \\
        ~& \citet{bevil-spring}           
        & 83.8  & 84.4 & \textbf{90.2} & 90.6 & 74.4 & 84.3 & 86.1 & 70.8 & 79.6 \\
        ~& HCL (Ours) & \textbf{84.3}            & \textbf{85.0} & \textbf{90.2} & \textbf{91.6}  & 75.9 &  84.0  &  \textbf{87.7}  &  \textbf{74.5} & \textbf{83.2} \\
        \midrule
        \multirow{5}{*}{\rotatebox[origin=c]{90}{{AMR3.0}}}& \citet{cai2020amr} & 78.0 & 78.5 & 88.5 & 83.7 & 68.9 & 75.7  & 81.9 & 63.7 & 73.2 \\
        ~& \citet{cai2020amr}$^G$ & 76.7 & 77.2 & 86.5 & 74.7 & 72.6 & 77.3 & 80.6 & 62.6 & 72.2 \\
        ~& \citet{zhou2021amr}
        & 80.3 & - & - & - & - & - & - & - & - \\
        ~& \citet{bevil-spring}
        & 83.0 & 83.5 & \textbf{89.8} & 87.2 & \textbf{73.0} & \textbf{82.7} & 85.4 & 70.4 & 78.9 \\
          ~& HCL (Ours) & \textbf{83.7} & \textbf{84.2} & 89.5 & \textbf{89.0} & \textbf{73.0} & 82.6 & \textbf{86.9} & \textbf{73.9} &  \textbf{82.4} \\
        \bottomrule
    \end{tabular}
    \caption{\textsc{Smatch} and fine-grained F1 scores on the AMR 2.0 and 3.0 test set. Our results are the average of 3 runs with different random seeds. Models$^G$ indicate models with graph re-categorization (a data processing method that may hurt the model generalization ability \citet{bevil-spring}).}
    \label{tab:main_results2.0}
\end{table*}

Motivated by \textit{learning core concepts first},
we propose Structure-level Curriculum (SC).
AMR graphs are organized in a hierarchy where the core semantics stay closely to the root \cite{cai-top2down},
thus SC divides all AMR sub-graphs into $N$ buckets according to their depths $\{S_i : i=1,2,...,N\}$,
where $S_i$ contains AMR sub-graphs with the depth $i$.
As shown in Figure \ref{fig:overview}(a), SC has $N$ training episodes, and each episode consists of $T_{sc}$ steps. 
In each step of the $i$-th episode, the training scheduler samples a batch of examples from buckets $\{S_j: j \le i\}$ to train the model.
When parsing a sentence into a sub-graph with the depth $d$, we append a special string ``parse to \textit{d} layers'' to the input sentence, and replace the start token of the decoder with an artificial token \textit{<d>}, so the model can perceive layers that need to be parsed.

\subsection{Instance-level Curriculum}
Inspired by \textit{learning easy instances first},
we propose Instance-level Curriculum (IC).
Figure \ref{fig:smatch-layer} shows AMR graphs with deeper layers can be regarded as harder instances for the flat pretrained model,
thus IC divides all AMR graphs into $M$ buckets according to their depths $\{I_i : i=1,...,M\}$,
where $I_i$ contains AMR graphs with the depth $i$.
As shown in Figure \ref{fig:overview}(b), IC has $M$ training episodes, and each episode consists of $T_{ic}$ steps. 
In each step of the $i$-th episode, the training scheduler samples a batch of examples from buckets $\{I_j: j \le i\}$ to train the model.
Specifically, we first use SC and then IC to train the model, since SC (follows learning core semantics first) is for AMR sub-graphs, which can be regarded as a warming-up stage of IC (obeys learning easy instances first), which is for AMR full graphs.

\section{Experiments}

\paragraph{Datasets and Evaluation Metrics}
We evaluate our hierarchical curriculum learning framework on two popular AMR benchmarks, AMR2.0 (LDC2017T10) and AMR3.0 (LDC2020T02). Please refer to the Appendix \ref{apx:datasets} for details of two benchmarks.
Following \citet{bevil-spring}, we use the \textsc{smatch} scores \cite{cai-smatch} and the fine-grained evaluation metrics \cite{dam-smatch-incremental}\footnote{https://github.com/mdtux89/amr-evaluation} to evaluate the performances.

\paragraph{Experiment Setups}
Our implementation is based on Huggingface's transformers
library \cite{wolf-etal-2020-transformers} and the open codebase of \citet{bevil-spring}\footnote{https://github.com/SapienzaNLP/spring}.
We use BART-large as our sequence-to-sequence model the same as \citet{bevil-spring}.
We utilizes RAdam \cite{liu-RAdam} as our optimizer with the learning rate $3$e-$5$.
The batch size is $2048$ graph linearization tokens with the gradient accumulation $10$.
Dropout is set to $0.25$ and beam size is $5$.   
The training steps $T_{sc}$ is $1000$ and $T_{ic}$ is $500$.
After the curriculum training, 
the model is trained for $30$ epochs on the training set.
We use cross-entropy as our loss function.
We train our model on a single NVIDIA TESLA V$100$ GPU with $32$GB memory.
We adopt the same post-processing process as \citet{bevil-spring}. Our code and model are available at \url{https://github.com/Wangpeiyi9979/HCL-Text2AMR}.

\paragraph{Main Results} 
We compare our method with previous approaches in Table \ref{tab:main_results2.0}.
As is shown, on AMR2.0 and AMR3.0, our hierarchical curriculum learning model achieves $84.3\pm0.1$ and $83.7\pm0.1$ \textsc{smatch} scores, and outperforms  \citet{bevil-spring} $0.5$ and $0.7$ \textsc{smatch} scores, respectively.
For the fine-grained results, our model achieves the best performance in $6$ out of $8$ metrics on both AMR2.0 and AMR3.0, which shows the effectiveness of our method.
Although \citet{cai2020amr} outperforms our model in Neg. and Wiki. on AMR2.0, they adopt a complex process, which may hurt the model generalization ability.
\citet{bevil-spring} outperforms slightly our model in Conc. and Wiki. on AMR3.0.
However, these metrics are unrelated to the AMR structure that our HCL focuses on.

\section{Analysis}
\begin{table}[t]
\centering
\small
    \begin{tabular}{lcc} \toprule
    Model & \textsc{AMR2.0} & \textsc{AMR3.0} \\
    \midrule
    Ours  &                   84.3  &    83.7     \\
    w/o instance curriculum & 84.1   &    83.5   \\ 
    w/o structure curriculum   & 84.0  &    83.3     \\
    w/o curricula  &  83.8  &    83.0 \\
    \bottomrule  
    \end{tabular}
\caption{The effect of our proposed curricula on the test set of AMR2.0 and AMR3.0. `w/o' denotes without.}
\label{tab:abla-study}
\end{table}

\paragraph{Structure Benefit}
In order to explore the effectiveness of our HCL framework for the structured AMR parsing.
We divide the fine-grained F1 scores into $2$ categories, ``structure-dependent'' (unlabelled, re-entrancy and SRL) and ``structure-independen'' (the left $5$ metrics).
Please refer to Appendix \ref{apx:metric division} for the reason for this division.
As shown in Table \ref{tab:main_results2.0}, 
compared with \citet{bevil-spring} (also a sequence-to-sequence model based on BART-large), our method achieves $2.97$ and $2.83$ average F1 scores improvement on $3$ structure-dependent metrics on AMR2.0 and AMR3.0, respectively, which proves HCL helps the flat sequence-to-sequence model better adapt to AMR with the hierarchical and complex structure.

\paragraph{Hard Instances Benefit}

\begin{figure}[t]
    \centering
    \includegraphics[width=1\linewidth]{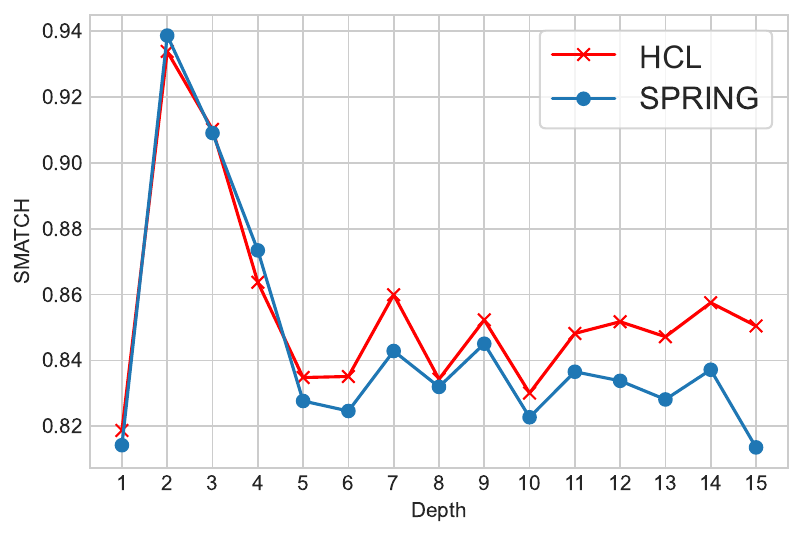}
    \caption{The average \textsc{smatch} scores of our HCL and \citet{bevil-spring} (SPRING) at different depths on the AMR2.0 test set.}
    \label{fig:smatch-layer-mix}
\end{figure}

Figure \ref{fig:smatch-layer-mix} shows the performances of our HCL and \citet{bevil-spring} (SPRING) at different layers.
As is shown, as the number of layers increases, HCL exceeds SPRING greater, which shows our HCL helps the model better handle hard instances.\footnote{An intuitive case study for the hard instance parsing is included in Appendix \ref{apx:case study}.}
In addition, to some extend, out-of-distribution (OOD) instances can be regarded as hard instances, thus we also consider the OOD situation.
\citet{bevil-spring} propose the OOD evaluation for AMR parsers.
Following \citet{bevil-spring}, we train our model on the training dataset of AMR2.0, and then evaluate it on 3 OOD test datasets, BIO, TLP and News3. Please refer to Appendix \ref{apx:datasets} for details of OOD datasets.
As shown in Table \ref{tab:ood}, our method outperforms \citet{bevil-spring} on all 3 OOD datasets, which shows our HCL framework can also improve the generalization ability of the model.

\paragraph{Ablation Study}

To illustrate the effect of our proposed curricula.
We conduct ablation studies by removing one curriculum at a time.
Table \ref{tab:abla-study} shows the \textsc{smatch} scores on both AMR2.0 and AMR3.0. 
As shown in Table \ref{tab:abla-study}, we can see both curricula are conducive to the performance of the model, and they are complementary to each other.
Specifically, the structure-level curriculum (SC) is more effective than the instance-level curriculum (IC).
We think the reason is that SC constructs AMR sub-graphs for training, which enhances the model's ability to perceive the AMR hierarchy.
\begin{table}[t]
    \centering
    \footnotesize
        \begin{tabular}{llccc}
                \toprule
                  Model&BIO & TLP &News3 \\
                \midrule
             \citet{bevil-spring} & 59.7 & 77.3 & 73.7\\
                 HCL (Ours) &\textbf{61.1} & \textbf{78.2} & \textbf{75.3} \\
                \bottomrule
            \end{tabular}
    \caption{Results on out-of-distribution data.}
    \label{tab:ood}
\end{table}

\section{Conclusion}
In this paper, we propose a Hierarchical Curriculum Learning (HCL) framework for sequence-to-sequence AMR parsing, which consists of Structure-level Curriculum (SC) and Instance-level Curriculum (IC).
inspired by human cognition,
SC follows the principle of learning the core concepts of AMR first, and
IC obeys the rule of learning easy instances first.
SC and IC train the model on different hierarchies (AMR sub-graphs and AMR full graphs).
Extensive experiments on AMR2.0, AMR3.0, structure-complex and out-of-distribution situations verify the effectiveness of HCL.

\section*{Acknowledgement}
The authors would like to thank the anonymous reviewers for their thoughtful and constructive comments, and \citet{bevil-spring} for their high-quality open codebase.
This paper is supported by the National Key Research and Development Program of China under Grant No. 2020AAA0106700, the National Science Foundation of China under Grant No.61936012 and 61876004, and NSFC project U19A2065.

\bibliography{HCL}
\bibliographystyle{acl_natbib}

\clearpage
\appendix
\section{Linearzation}
\begin{figure}[h]
    \centering
    \includegraphics[width=0.85\linewidth]{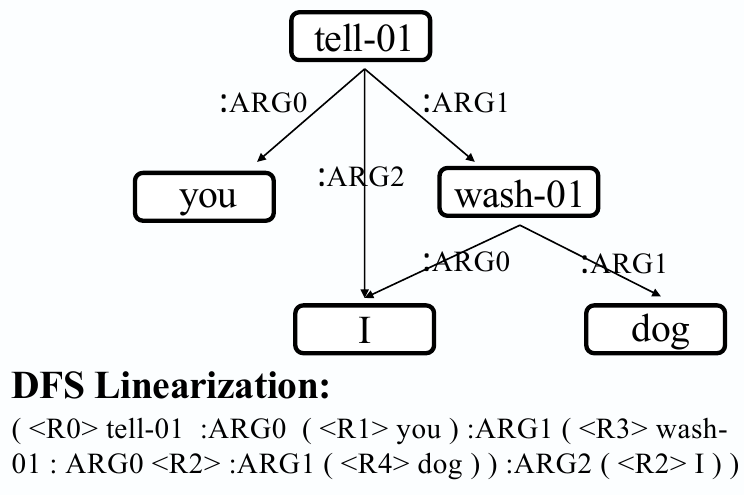}
    \caption{The linearization for the AMR graph of the sentence ``You told me to wash the dog''.}
    \label{fig:linearzation}
\end{figure}
\label{apx:linearzation}
As shown in Figure \ref{fig:linearzation}, following \citet{bevil-spring}, the AMR graph is linearized by the DFS-based linearization method according to the edge order (`:ARG0'$\to$`:ARG1'$\to$`:ARG2'). Variables of the AMR graph are set to a series of special tokens \textit{<R0>, <R1>, <R2>, <R3>, <R4>}, and the depth is marked by parentheses.

\section{Datasets}
\label{apx:datasets}
\subsection{In-domain Distribution}
\paragraph{AMR2.0} (LDC2017T10) contains $36,521$, $1,368$ and $1,371$ sentence-AMR pairs in training, development and testing sets, respectively.
\paragraph{AMR3.0} (LDC2020T02) is larger than AMR2.0 in size, which contains $55,635$, $1,722$ and $1,898$ sentence-AMR pairs for training development and testing set, respectively. AMR3.0 is a superset of AMR2.0.
\subsection{Out-domain Distribution}
\paragraph{BIO} is a test set of the Bio-AMR corpus, consisting of $500$ instances.
\paragraph{TLP} is a AMR dataset annoated on the children's novel \textit{The Little Prince} (version 3.0), consisting of $1,562$ instances.
\paragraph{New3} is a sub-set of AMR3.0, which is not included in the AMR2.0 training set, consisting of $527$ instances.

\section{Fine-grained Metric Division}
\label{apx:metric division}
There are $8$ fine-grained AMR metrics:
(1) \textbf{Unlabeled}: Smatch score computed on the predicted graphs after removing all edge labels.
(2) \textbf{No WSD.}: Smatch score while ignoring Propbank senses (e.g., duck-01 vs duck-02).
(3) \textbf{Named Ent.}: F-score on the named entity recognition (:name roles).
(4) \textbf{Wikification}: F-score on the wikification (:wiki roles).
(5) \textbf{Negation}: F-score on the negation detection (:polarity roles).
(6) \textbf{Concepts}: F-score on the concept identification task.
(7) \textbf{Reentrancy}: Smatch computed on reentrant edges only, e.g., the edges of node `I' in Figure \ref{apx:linearzation}.
(8) \textbf{SRL}: Smatch computed on :ARG-i roles only.

We only regard Unlabeled, Reentrancy and SRL as ``structure-dependent'' metrics, since:
(1) Unlabeled does not consider any edge labels, and only considers the graph structure.
(2) Reentrancy is a typical structure feature for the AMR graph. Without reentrant edges, the AMR graph is reduced to a tree.
(3) SRL denotes the core-semantic relation of the AMR, which determines the core structure of the AMR.
(4) As described above, all other metrics have little relationship with the structure.

\section{Case Study}
\label{apx:case study}
\begin{figure}[h]
    \centering
    \includegraphics[width=0.95\linewidth]{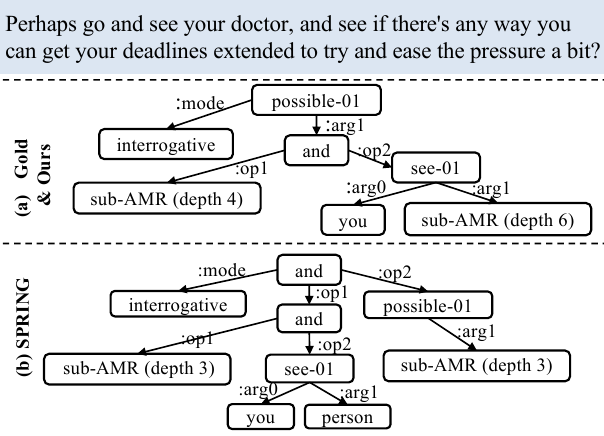}
    \caption{A specific case from the test set of AMR2.0. For the input sentence, our method achieves the right AMR, while the baseline model (i.e., SPRING \cite{bevil-spring}) gets a shallower and wrong structure AMR. }
    \label{fig:case study}
\end{figure}
Figure \ref{fig:case study} shows a case study (we omit some details of AMR graphs for a more clear description).
As is illustrated, our method achieves the right AMR for the input sentence.
However, the AMR parsed by the SPRING model (depth:5) is shallower than the gold AMR (depth:9), and their structures are also different (e.g., the root of the gold AMR and the SPRING parsed AMR are  `possible-01' and `and', respectively).
This case intuitively shows our HCL framework can help the model better handle the hard instance with complex structure.

\end{document}